\documentclass[a4paper,twoside]{article}

\usepackage{epsfig}
\usepackage{subcaption}
\usepackage{calc}
\usepackage{amssymb}
\usepackage{natbib} 
\usepackage{amstext}
\usepackage{amsmath}
\usepackage{amsthm}
\usepackage{multicol}
\usepackage{pslatex}
 \usepackage{url} 
\usepackage{algorithm2e}
\usepackage{hyperref}
\usepackage{booktabs}
\usepackage[bottom]{footmisc}
\usepackage{SCITEPRESS}     

\begin{document}

\title{Divergent Emotional Patterns in Disinformation on Social Media? \\ An Analysis of Tweets and TikToks about the DANA in Valencia}

\author{
  \authorname{
    Iván Arcos\sup{1}, Paolo Rosso\sup{1,2}, Ramón Salaverría\sup{3}
  }
  \affiliation{
    \sup{1}PRHLT Research Center, Universitat Politècnica de València, Valencia, Spain \\
    \sup{2}ValgrAI Valencian Graduate School and Research Network of Artificial Intelligence, Spain \\
    \sup{3}School of Communication, Universidad de Navarra, Pamplona, Spain
  }
  \email{iarcgab@etsinf.upv.es, prosso@dsic.upv.es, rsalaver@unav.es}
}


\abstract{
This study investigates the dissemination of disinformation on social media platforms during the DANA event (DANA is a Spanish acronym for \textit{Depresión Aislada en Niveles Altos}, translating to high-altitude isolated depression) that resulted in extremely heavy rainfall and devastating floods in Valencia, Spain, on October 29, 2024. We created a novel dataset of 650 TikTok and X posts, which was manually annotated to differentiate between disinformation and trustworthy content. Additionally, a Few-Shot annotation approach with GPT-4o achieved substantial agreement (Cohen's kappa of 0.684) with manual labels. Emotion analysis revealed that disinformation on X is mainly associated with increased sadness and fear, while on TikTok, it correlates with higher levels of anger and disgust. Linguistic analysis using the LIWC dictionary showed that trustworthy content utilizes more articulate and factual language, whereas disinformation employs negations, perceptual words, and personal anecdotes to appear credible. Audio analysis of TikTok posts highlighted distinct patterns: trustworthy audios featured brighter tones and robotic or monotone narration, promoting clarity and credibility, while disinformation audios leveraged tonal variation, emotional depth, and manipulative musical elements to amplify engagement. In detection models, SVM+TF-IDF achieved the highest F1-Score, excelling with limited data. Incorporating audio features into \textit{roberta-large-bne} improved both Accuracy and F1-Score, surpassing its text-only counterpart and SVM in Accuracy. GPT-4o Few-Shot also performed well, showcasing the potential of large language models for automated disinformation detection.  These findings demonstrate the importance of leveraging both textual and audio features for improved disinformation detection on multimodal platforms like TikTok.}

\onecolumn \maketitle \normalsize \setcounter{footnote}{0} \vfill

\section{\uppercase{Introduction}}
\label{sec:introduction}

In the digital era, social media platforms such as X (formerly Twitter) and TikTok revolutionized communication and information dissemination. These platforms facilitate instant sharing and broad interactions, enabling users to engage with content in real-time and reach vast audiences with unprecedented speed. While this democratization of information access offers numerous benefits, it also presents significant challenges, particularly concerning the spread of false or misleading information.

\begin{figure}[ht]
    \centering
    \includegraphics[width=1\linewidth]{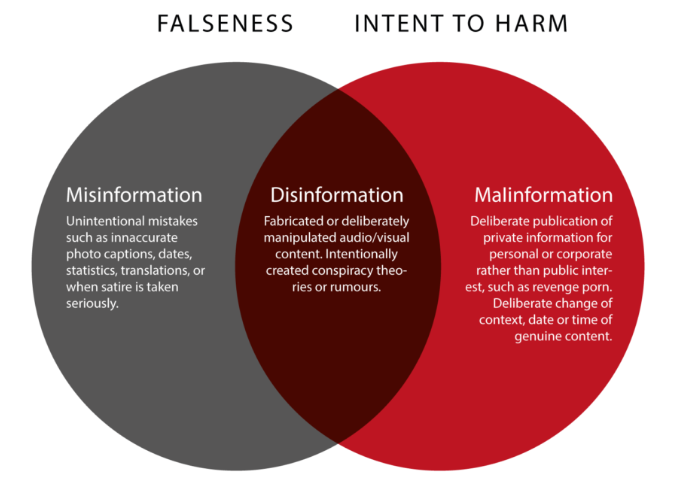} 
    \caption[Relationship among Disinformation, Misinformation, and Malinformation]{Relationship among Disinformation, Misinformation, and Malinformation.\citep{wardle2017information}\footnotemark}
    \label{fig:conceptos}
\end{figure}

\footnotetext{\url{https://firstdraftnews.org/latest/coe_infodisorder/}}

Disinformation, defined by \cite{european2019tackling}  as  “verifiably  false  or  misleading information  created,  presented,  and  disseminated  for  financial  gain  or  intentional  deception  of  the public”, has become a pervasive issue on social media. Unlike misinformation, which involves the unintentional sharing of incorrect information without malicious intent, disinformation is strategically crafted to manipulate public perception, influence opinions, and achieve specific agendas. Additionally, malinformation refers to the dissemination of genuine information with the intent to harm or manipulate, further complicating the information ecosystem. These three categories are illustrated in Figure \ref{fig:conceptos}.

The design of social media platforms is characterized by algorithms that prioritize engagement, the formation of echo chambers that reinforce existing beliefs, and the ease of content sharing. These features facilitate the rapid and widespread propagation of disinformation. This environment allows malicious actors, including those whose aim is to spread disinformation to increase their number of followers, as well as automated accounts and coordinated misinformation campaigns, to effectively amplify false narratives. Furthermore, the viral nature of social media content means that disinformation can quickly reach and influence large segments of the population, often outpacing fact-checking and regulatory efforts aimed at curbing its spread.

Understanding the mechanisms behind the dissemination of disinformation is crucial for developing effective countermeasures. Emotional and linguistic patterns play a significant role in how disinformation resonates with audiences and spreads across platforms. Analyzing these patterns can provide insights into the narrative employed by those who create and distribute false information, as well as the emotional triggers that make such content more likely to be shared and believed \citep{mcloughlin2024misinformation}.

This study focuses on the spread of disinformation during the DANA event in Valencia, Spain, on October 29, 2024, which caused severe flooding, resulting in 224 deaths and three missing persons. By examining emotional and linguistic patterns in 650 annotated posts from TikTok and X, we aim to elucidate the strategies used in disseminating false narratives and their impact on public sentiment. The research employs both traditional machine learning models and advanced language models, such as transformers and Large Language Models (LLMs), to detect disinformation, highlighting the strengths and limitations of each approach.

The remainder of this article is organized as follows: Section \ref{sec:related_work} reviews related literature on disinformation detection in social media. Section \ref{sec:datasets} details the data collection and annotation processes. Section \ref{sec:emotion_analysis} presents the emotion analysis conducted on the dataset. Section \ref{sec:linguistic_analysis} explores the linguistic patterns using the LIWC dictionary. Section \ref{sec:audio_analysis} discusses the analysis of audio variables in TikTok posts. Section \ref{sec:models} describes the models used for disinformation detection and their performance. Finally, Section \ref{sec:conclusions} concludes the study and outlines directions for future research.

\section{Related Work}
\label{sec:related_work}

The spread of misinformation and disinformation on social media has garnered significant research attention due to its profound impact on public opinion, democratic processes, and public health \citep{wang2022disinformation}. The viral nature of social media platforms facilitates the swift dissemination of both true and false news \citep{vosoughi2018spread}.

A seminal study by \citet{vosoughi2018spread} analyzed the diffusion patterns of true and false news on X from 2006 to 2017, revealing that false news spreads significantly farther, faster, deeper, and more broadly than true news, particularly in political contexts. This enhanced virality is attributed to the novelty and emotional content of false stories, which evoke fear, disgust, and surprise, thereby increasing user engagement. Recent research has found that outrage is a key emotion that facilitates the spread of misinformation particularly on X \citep{mcloughlin2024misinformation}. Complementing these findings, \citet{allcott2017social} examined the role of fake news in the 2016 US presidential election, highlighting its substantial reach and influence, especially for stories favoring specific political candidates. The study underscored that users are more inclined to believe fake news that aligns with their preexisting beliefs, particularly within ideologically segregated social networks.

Addressing the challenges posed by misinformation, researchers have proposed various interventions. \citet{pennycook2019implied} introduced the concept of the implied truth effect, demonstrating that attaching warnings to a subset of fake news headlines can inadvertently increase the perceived accuracy of untagged false headlines, highlighting the complexity of designing effective misinformation mitigation strategies.

The COVID-19 pandemic further highlighted the dangers of misinformation, giving rise to an “infodemic.” \citet{islam2020covid} conducted a global social media analysis to assess the impact of false information related to COVID-19, finding that misinformation about treatments, transmission, and mortality rates fueled public panic and hindered effective public health responses.

In the realm of fake news detection, \citet{shu2017fake} provided a comprehensive review of text mining techniques employed to identify false news on social media, proposing a conceptual framework that leverages both content-based and context-based features to enhance detection accuracy.

Multimodal approaches utilize both textual and visual data to enhance fake news identification. For instance, \citet{rashid2024unraveling} introduced MuAFaNI, a framework that combines text and image representations using RoBERTa and ResNet-50 models to assess news authenticity. Similarly, \citet{qu2024qmfnd} proposed QMFND, a quantum multimodal fusion-based model that integrates image and textual features through a quantum convolutional neural network, thereby improving detection accuracy. Further advancing the field, \citet{lee2024emotional} explored the emotional effects of multimodal disinformation, demonstrating that video-based misinformation significantly increases anxiety and misperceptions compared to text-only content, highlighting the enhanced persuasive power of multimodal disinformation and its implications for public trust and decision-making.

Other notable contributions include \citet{kumar2024feature}, which developed a multimodal fake news detection model incorporating textual and visual features alongside feature engineering techniques to capture the behavior of fake news propagators. \citet{shan2024multimodal} employed similarity inference and adversarial networks to fuse textual and visual features, thereby enhancing detection accuracy on social media platforms. Additionally, \citet{pan2024mfae} and \citet{luvembe2024caf} introduced models that leverage graph convolutional networks and complementary attention mechanisms to align and fuse multimodal features, achieving significant accuracy improvements on benchmark datasets.

The emergence of deepfakes has further complicated the disinformation landscape. \citet{vaccari2020deepfakes} investigated the impact of synthetic political videos, revealing that deepfakes contribute to uncertainty and reduce trust in news sources, thereby undermining the integrity of public discourse.

While multimodal approaches represent a significant advancement by leveraging both textual and visual data, the application of multimodal techniques specifically to TikTok videos remains underexplored. This study aims to address this gap by focusing on the multimodal characteristics of TikTok content, initially leveraging transcriptions of video audio, with the intention to incorporate video and audio features in future work.

\section{\uppercase{Datasets}}
\label{sec:datasets}

\subsection{Data Extraction}
\label{subsec:data_annotation}

The datasets used in this study were extracted from TikTok and X using crawlers provided by \textit{Apify}\footnote{\url{https://console.apify.com/}}. For TikTok, the \textit{clockworks/tiktok-scraper}\footnote{\url{https://apify.com/clockworks/tiktok-scraper}} was employed, while for X, the \textit{fastcrawler/Tweet-Fast-Scraper}\footnote{\url{https://apify.com/fastcrawler/tweet-fast-scraper/api/javascript}} was utilized. Additionally, the datasets will be made public for research purposes to facilitate further studies and enhance transparency in disinformation research.

The data extraction process was conducted using keywords related to the DANA phenomenon and associated disinformation narratives. The keywords, summarized in Table~\ref{tab:keywords_translation}, were carefully selected to capture content that potentially spread disinformation or expressed emotional reactions during the DANA phenomenon.

\begin{table}[h!]
\centering
\caption{Keywords Used for Data Extraction and Their Translations}
\resizebox{1\columnwidth}{!}{
\begin{tabular}{|l|l|}
\hline
\textbf{Keyword (Spanish)} & \textbf{Translation (English)} \\ \hline
\textit{DANA conspiración} & \textit{DANA conspiracy} \\ \hline
\textit{DANA fallecidos ocultación} & \textit{DANA deceased concealment} \\ \hline
\textit{DANA engaño} & \textit{DANA deception} \\ \hline
\textit{DANA manipulación} & \textit{DANA manipulation} \\ \hline
\textit{DANA mentiras} & \textit{DANA lies} \\ \hline
\textit{DANA ataque climático} & \textit{DANA climate attack} \\ \hline
\textit{DANA manipular la verdad} & \textit{DANA manipulate the truth} \\ \hline
\textit{DANA desinformación} & \textit{DANA disinformation} \\ \hline
\textit{DANA falsedades} & \textit{DANA falsehoods} \\ \hline
\textit{DANA ocultación} & \textit{DANA concealment} \\ \hline
\textit{DANA Rubén Gisbert} & \textit{DANA Rubén Gisbert} \\ \hline
\textit{DANA Alvise Pérez} & \textit{DANA Alvise Pérez} \\ \hline
\textit{DANA Iker Jiménez} & \textit{DANA Iker Jiménez} \\ \hline
\textit{DANA Vito Quiles} & \textit{DANA Vito Quiles} \\ \hline
\textit{DANA Cruz Roja falsa ayuda} & \textit{DANA Cruz Roja fake aid} \\ \hline
\textit{DANA Bonaire cementerio} & \textit{DANA Bonaire cemetery} \\ \hline
\textit{DANA ayuda rechazada} & \textit{DANA rejected aid} \\ \hline
\textit{DANA pronóstico incorrecto} & \textit{DANA incorrect forecast} \\ \hline
\textit{DANA radar sin funcionar} & \textit{DANA radar not working} \\ \hline
\textit{DANA provocada presas} & \textit{DANA caused by dams} \\ \hline
\end{tabular}
}
\label{tab:keywords_translation}
\end{table}

The extraction focused on filtering posts that matched these keywords, aiming to collect content that potentially spread disinformation  or expressed emotional reactions during the DANA phenomenon.

For TikTok videos, the transcription of spoken content was extracted using the Whisper-medium model\footnote{\url{https://huggingface.co/openai/whisper-medium}}, which is designed for efficient and accurate speech-to-text transcription. Additionally, text appearing in the videos was extracted using PaddleOCR\footnote{\url{https://github.com/PaddlePaddle/PaddleOCR}}, a state-of-the-art Optical Character Recognition (OCR) tool. These methods ensured the comprehensive collection of both spoken and visual textual content for analysis.

\subsection{Bias Assessment}

To assess potential biases introduced by keyword-based data collection, we applied the Weirdness Index (WI) \citep{Poletto2020ResourcesAB}, which measures how much certain words in a dataset deviate from their expected frequencies in a reference corpus.

As a reference, we collected social media content using broad, non-targeted keywords: “DANA Valencia”. This corpus captures general discussions on the DANA phenomenon, including neutral and informative posts, rather than those explicitly linked to disinformation. By comparing our dataset against this reference, we evaluated whether our keyword selection introduced thematic biases.

Table~\ref{tab:weirdness_results} presents the WI statistics for both datasets. A high WI indicates that certain words are disproportionately represented compared to the reference corpus.

\begin{table}[h]
    \centering
    \begin{tabular}{|l|c|c|}
        \hline
        \textbf{Metric} & \textbf{Tweets} & \textbf{TikToks} \\
        \hline
        WI Global (Mean) & 4.86 & 1.76 \\
        WI Global (Median) & 2.37 & 0.79 \\
        WI Standard Deviation & 45.63 & 3.23 \\
        Words with WI $>$ 2 & 55.64\% & 24.90\% \\
        \hline
    \end{tabular}
    \caption{Weirdness Index (WI) results comparing datasets.}
    \label{tab:weirdness_results}
\end{table}

The tweets dataset shows a significantly higher WI, indicating that keyword filtering has amplified specific narratives. Given the constraints in data availability and manual annotation time, our objective was to capture a broad range of disinformation-related themes rather than general discussions on the DANA phenomenon. The keyword selection was designed to encompass various disinformation topics, including conspiracy theories, exaggerated impacts, and misleading emergency response claims.  

In contrast, the TikTok dataset exhibits a lower WI, suggesting a more diverse linguistic distribution. Although the same keyword selection strategy was used, TikTok’s transcription-based content structure results in a dataset that, while still containing disinformation, is less narrowly focused compared to tweets.

\subsection{Data Annotation}

The collected TikToks and tweets were further annotated to create a labeled subset for analysis. Each post was examined to determine its relation to the DANA phenomenon and the spread of disinformation. Fact-checking was conducted using the reputable websites \textit{Maldita}\footnote{\url{https://maldita.es}}, \textit{Newtral}\footnote{\url{https://newtral.es}}, and \textit{EFE Verifica}\footnote{\url{https://verifica.efe.com/}}.

The annotation process involved assigning one of two labels to each post:
\begin{itemize}
    \item \textbf{Label 0}. Assigned to posts that discuss the DANA phenomenon without spreading disinformation  or falsehoods. This category also includes posts that actively report or debunk disinformation .
    \item \textbf{Label 1}. Assigned to posts that directly spread disinformation  or falsehoods about the DANA phenomenon.
\end{itemize}

Table~\ref{tab:label_distribution} shows the distribution of labels for TikToks and tweets. Efforts were made to ensure a balanced representation between the two classes (0 and 1) to facilitate unbiased analysis. 

\begin{table}[h!]
\centering
\caption{Label distribution for TikToks and Tweets.}
\begin{tabular}{|c|c|c|c|}
\hline
\textbf{Label} & \textbf{TikToks} & \textbf{Tweets} & \textbf{Total} \\ \hline
0 (Trustworthy) & 131 & 177 & 308 \\ \hline
1 (Disinformation) & 137 & 205 & 342 \\ \hline
\textbf{Total} & \textbf{268} & \textbf{382} & \textbf{650} \\ \hline
\end{tabular}
\label{tab:label_distribution}
\end{table}

To better illustrate the difference between disinformation and trustworthy content, Table~\ref{tab:examples_tweets} presents one example of each category. These examples highlight the type of content analyzed and the criteria used for labeling. 

\begin{figure}[h!]
    \centering
    \begin{subfigure}[b]{\columnwidth}
        \centering
        \includegraphics[width=\columnwidth]{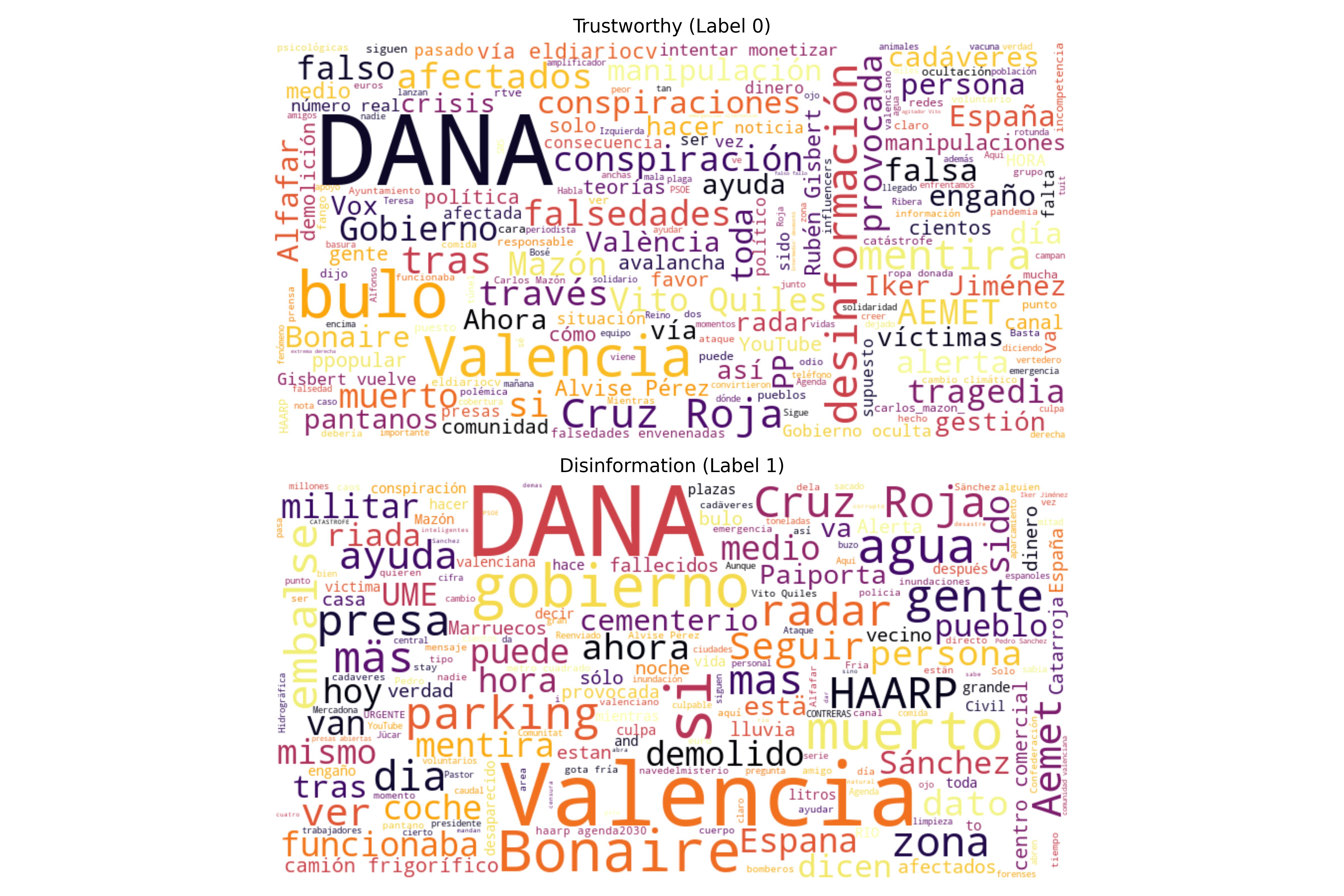}
        \caption{Word Clouds for X: Disinformation and Trustworthy Posts}
        \label{fig:wc_twitter}
    \end{subfigure}
    \vfill
    \begin{subfigure}[b]{\columnwidth}
        \centering
        \includegraphics[width=\columnwidth]{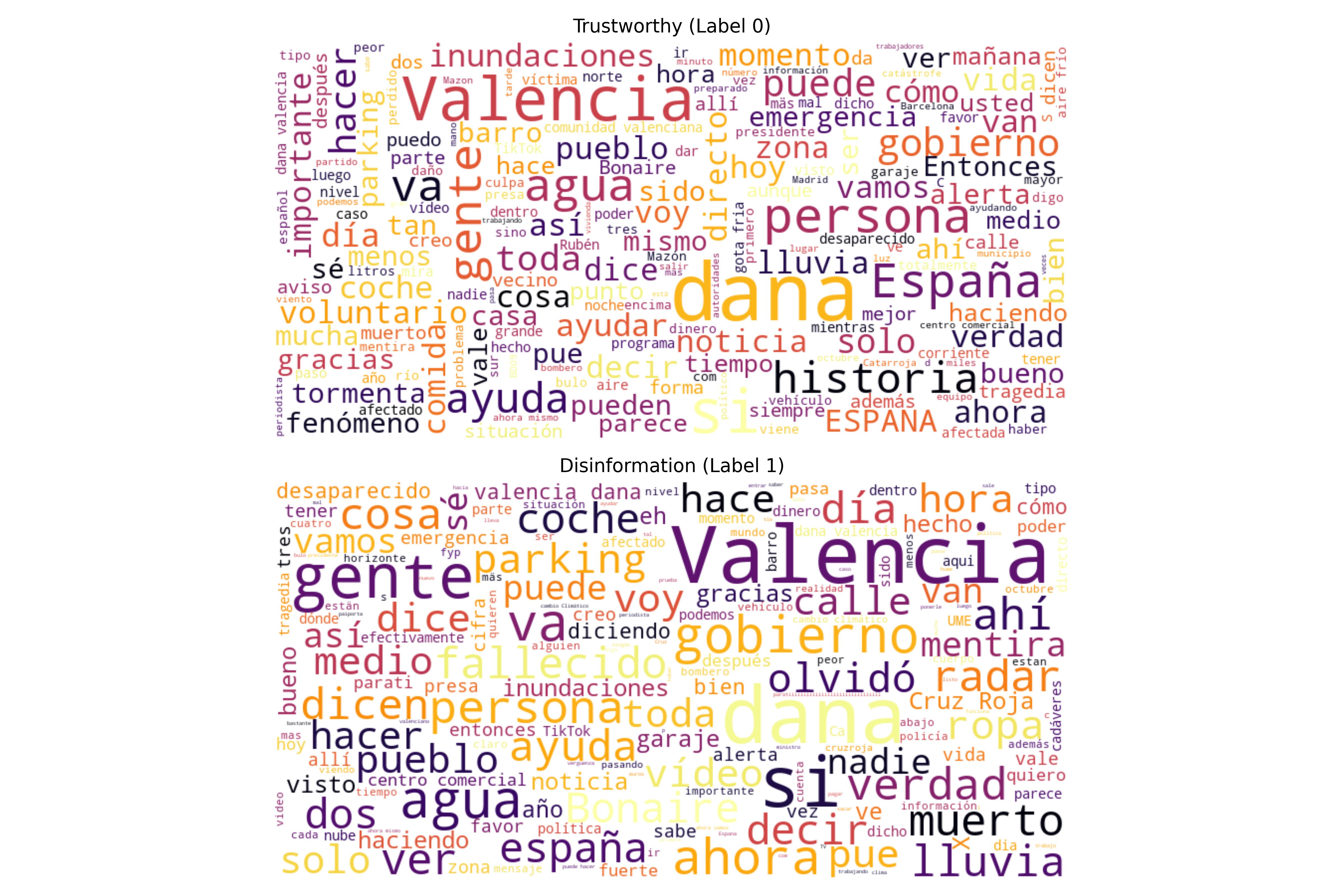}
        \caption{Word Clouds for TikTok: Disinformation and Trustworthy Posts}
        \label{fig:wc_tiktok}
    \end{subfigure}
    \caption{Word clouds from X and TikTok.}
    \label{fig:wordclouds}
\end{figure}

The word clouds in Figure~\ref{fig:wordclouds} highlight that the vocabulary across both social networks is strikingly similar. On X, terms related to \textit{trustworthy} often include \textit{bulo} (hoax), \textit{tragedia} (tragedy), \textit{falso} (false), and \textit{desinformación} (disinformation), alongside mentions of public figures such as Iker Jiménez\footnote{Iker Jiménez is a journalist.}, Vito Quiles\footnote{Vito Quiles is a pseudojournalist.}, Rubén Gisbert\footnote{Rubén Gisbert is a pseudojournalist.}, and Alvise Pérez\footnote{Alvise Pérez is the leader of \textit{Se Acabó La Fiesta} (\textit{The Party is Over}).}. These terms often appear in posts criticizing or exposing falsehoods. In \textit{disinformation}, words such as \textit{embalse} (reservoir), \textit{presa} (dam), \textit{HAARP}\footnote{Alleged climate attacks generated from HAARP (High-Frequency Active Auroral Research Program).}, \textit{Sánchez} (Spanish Prime Minister Pedro Sánchez), \textit{UME}\footnote{UME (Unidad Militar de Emergencias) is the Spanish Military Emergencies Unit.}, \textit{demolido} (demolished), \textit{cementerio} (graveyard), and \textit{AEMET}\footnote{AEMET (Agencia Estatal de Meteorología) is the Spanish national meteorological agency.} are frequently observed.

On TikTok, the focus of terms associated with \textit{trustworthy} shifts towards community and natural disaster responses, with words such as \textit{inundaciones} (floods), \textit{voluntario} (volunteer), \textit{fenómeno} (phenomenon), \textit{tormenta} (storm), \textit{ayudar} (help), and \textit{zona} (area) standing out. Disinformation in this context is more closely linked to terms like \textit{Cruz Roja} (Red Cross), \textit{ropa} (clothing), \textit{verdad} (truth), \textit{gobierno} (government), \textit{radar} (radar), and \textit{cadáveres} (corpses).

\begin{table}[h!]
\centering
\caption{Examples of Disinformation and Trustworthy Tweets (and their translation into English).}
\begin{tabular}{|p{0.45\textwidth}|}
\hline
\small \textbf{Trustworthy Tweet (Label 0)} \\ \hline
\small \textit{Miles de voluntarios se movilizan desde Les Arts hacia las zonas afectadas. Matías Prats informa que no hay víctimas mortales en el túnel.} \\ 
\small (Thousands of volunteers are mobilizing from Les Arts towards the affected areas. Matías Prats reports that there are no mortal victims in the tunnel.) \\ \hline
\small \textbf{Disinformation Tweet (Label 1)} \\ \hline
\small \textit{Me acaban de comunicar que ya han sacado 86 fallecidos del parking de Bonaire. El aparcamiento del centro comercial tiene 5700 plazas.} \\ 
\small (I have just been informed that 86 deceased have already been removed from the Bonaire parking lot. The shopping center's parking lot has 5700 spaces.) \\ \hline
\end{tabular}
\label{tab:examples_tweets}
\end{table}

\subsection{Few-Shot Annotation with GPT-4o}

A Few-Shot annotation approach was tested using GPT-4o, a variant of GPT-4 optimized for enhanced efficiency and performance. While all annotations in this study were conducted manually, GPT-4o was applied to a subset of tweets to explore the potential of large language models for automated annotation. 

 Although the primary focus of our analysis was on two main categories (\textit{Trustworthy information} and \textit{Disinformation}), we discovered that refining the prompt to include four detailed categories and then collapsing them into two yielded higher agreement values between GPT-4o and manual annotations. This strategy allowed for a more nuanced understanding of the tweets during the initial annotation process, which improved the precision and reliability of the final binary labels.

\subsubsection{Prompt Design for GPT-4o}

To classify the tweets into predefined categories, the following clean and structured prompt was provided to GPT-4o. For readers interested in the full details of the prompt structure, please refer to \textbf{Appendix \ref{sec:appendix_prompt}}.

\subsubsection{Results and Evaluation}

Table~\ref{tab:annotation_agreement_4} shows the annotation agreement matrix for the four predefined categories, where GPT-4o achieved a Cohen's kappa of 0.684, indicating a substantial agreement with manual annotations. This highlights the effectiveness of GPT-4o in capturing nuanced differences between the categories.

\begin{table}[h!]
\centering
\caption{Annotation Agreement for Four Categories.}
\resizebox{0.6\columnwidth}{!}{ 
\begin{scriptsize} 
\begin{tabular}{|l|c|c|c|c|}
\hline
\textbf{} & \textbf{0} & \textbf{1} & \textbf{2} & \textbf{3} \\ \hline
\textbf{0} & 15 & 1 & 0 & 2 \\ \hline
\textbf{1} & 1 & 13 & 1 & 1 \\ \hline
\textbf{2} & 3 & 4 & 28 & 5 \\ \hline
\textbf{3} & 4 & 2 & 3 & 36 \\ \hline
\end{tabular}
\end{scriptsize}
}
\label{tab:annotation_agreement_4}
\end{table}

When the categories were reduced to two groups, \textit{Trustworthy (0)} and \textit{Disinformation (1)}, the Cohen's kappa further improved to 0.695 (Table~\ref{tab:annotation_agreement_2}). 

\begin{table}[h!]
\centering
\caption{Annotation Agreement for Two Categories.}
\resizebox{0.4\columnwidth}{!}{ 
\begin{scriptsize} 
\begin{tabular}{|l|c|c|}
\hline
\textbf{} & \textbf{0} & \textbf{1} \\ \hline
\textbf{0} & 66 & 8 \\ \hline
\textbf{1} & 9 & 36 \\ \hline
\end{tabular}
\end{scriptsize}
}
\label{tab:annotation_agreement_2}
\end{table}

These results demonstrate the power of GPT-4o, achieving substantial agreement with manual annotations. Such capabilities can significantly reduce the time and effort required for annotating large-scale datasets, particularly in the context of disinformation  detection during events like DANA.

\section{\uppercase{Emotion Analysis in Tweets and TikToks}}
\label{sec:emotion_analysis}
This study analyzed the emotions present in tweets and TikToks using the \textit{twitter-xlm-roberta-emotion-es} model\footnote{\url{https://huggingface.co/daveni/twitter-xlm-roberta-emotion-es}}. The model identifies six emotions: \textit{sadness}, \textit{joy}, \textit{anger}, \textit{surprise}, \textit{disgust}, \textit{fear}, and \textit{others}. These categories align with the basic emotions proposed by Paul Ekman, which are considered universal across cultures and rooted in human evolution \cite{Ekman1992}. Non-parametric tests (Mann–Whitney U) were performed to determine emotional patterns in disinformation across both X and TikTok.

\subsection{Emotional Patterns in X}

Table~\ref{tab:twitter_emotions} presents the results for emotional analysis in tweets. Notably, tweets with disinformation showed significantly higher sadness (\textit{p} = 0.000061). This often corresponds to claims exaggerating the number of casualties, e.g., “El engaño sobre el número de muertos es cruel, indignante e inhumano” (“The deception about the number of deaths is cruel, outrageous, and inhumane”).

Disinformation also showed higher fear levels (\textit{p} = 0.031), linked to claims about catastrophic events, such as dam failures, e.g., “Lo que pasa es que el embalse de Tous está al 97\%. Si abren las compuertas, viene una riada” (“What’s happening is that the Tous reservoir is at 97\%. If they open the gates, a flood is coming”).

\begin{table}[h!]
\centering
\caption{Emotional patterns in X.}
\resizebox{1\columnwidth}{!}{
\begin{tabular}{|l|c|c|c|c|c|}
\hline
\textbf{Emotion} & \textbf{Mean (0)} & \textbf{Std (0)} & \textbf{Mean (1)} & \textbf{Std (1)} & \textbf{\textit{p}-value} \\ \hline
\textbf{Sadness} & 0.077 & 0.131 & 0.176 & 0.235 & 0.000061 \\ \hline
\textbf{Fear} & 0.012 & 0.007 & 0.014 & 0.008 & 0.031030 \\ \hline
\textbf{Others} & 0.622 & 0.296 & 0.522 & 0.309 & 0.002906 \\ \hline
\end{tabular}
}
\label{tab:twitter_emotions}
\end{table}

\subsection{Emotional Patterns in TikToks}
Table~\ref{tab:tiktok_emotions} shows emotional patterns in TikToks. Unlike X, sadness is more prevalent in TikToks without disinformation (\textit{p} = 0.000380), often focusing on the tragedy itself, e.g., “Lamentablemente las grandes inundaciones que se están provocando en Valencia España han dejado hasta la fecha más de 2500 desaparecidos” (“Unfortunately, the severe floods happening in Valencia, Spain, have left over 2,500 people missing so far”).

Disinformation is associated with higher levels of anger (\textit{p} = 0.000003) and disgust (\textit{p} = 0.000003), reflecting frustration or exaggeration, e.g., “Las ayudas de los 6.000 euros para los afectados por la DANA las tienes que devolver en tres meses” (“The €6,000 aid for those affected by DANA must be repaid within three months”), or “Luego digan que no hay muertos en las calles, señor. Que no hay muertos en las calles. Acaban de tapar uno aquí con una toalla. Después de cuatro días recogiendo y amontonando cosas entre la mierda. ¿Eh? Y no había muertos. ¿Eh? No habían muertos” (“Then they say there are no dead bodies in the streets, sir. That there are no dead bodies in the streets. They just covered one here with a towel. After four days of picking up and piling things among the filth. Huh? And there were no dead bodies. Huh? There were no dead bodies”).

\begin{table}[h!]
\centering
\caption{Emotional patterns in TikToks.}
\resizebox{1\columnwidth}{!}{
\begin{tabular}{|l|c|c|c|c|c|}
\hline
\textbf{Emotion} & \textbf{Mean (0)} & \textbf{Std (0)} & \textbf{Mean (1)} & \textbf{Std (1)} & \textbf{\textit{p}-value} \\ \hline
\textbf{Sadness} & 0.294 & 0.196 & 0.218 & 0.191 & 0.000380 \\ \hline
\textbf{Anger} & 0.165 & 0.143 & 0.262 & 0.187 & 0.000003 \\ \hline
\textbf{Disgust} & 0.027 & 0.018 & 0.040 & 0.025 & 0.000003 \\ \hline
\end{tabular}
}
\label{tab:tiktok_emotions}
\end{table}

\section{\uppercase{Linguistic Analysis Using LIWC}}
\label{sec:linguistic_analysis}

In addition to emotional analysis, linguistic patterns were examined using the LIWC dictionary \cite{liwc}\footnote{\url{https://www.liwc.app/}}, which groups words into dimensions such as \textit{Linguistic Dimensions}, \textit{Psychological Processes}, \textit{Personal Concerns}, and \textit{Spoken Categories}.

\subsection{Linguistic Patterns in TikToks}

Table~\ref{tab:tiktok_liwc} summarizes the linguistic patterns in TikToks. When disinformation is not present, the discourse tends to be more articulate, with a higher use of prepositions and terms related to health, often informing about infection risks or providing warnings e.g., “El riesgo de infección en las zonas afectadas por la DANA sigue siendo muy alto debido a las aguas estancadas y contaminadas...” (“The risk of infection in areas affected by the DANA remains very high due to stagnant and contaminated water...”). Such videos also include more references to time and specific events e.g., “¡Atención! Una nueva DANA traerá humedad y precipitaciones a partir del martes por la noche” (“Warning! A new DANA will bring humidity and rainfall starting Tuesday night”).

Conversely, videos spreading disinformation often use more negation and profanities (Maldec), reflecting frustration or skepticism. These videos also employ more perceptual words (Percept) to suggest firsthand observations or unverifiable claims e.g., “En las noticias no sale toda la verdad, están desaguando el centro comercial de aquí dicen que hay más de 800 me ha dicho un guardia civil...” (“The news isn’t telling the whole truth; they’re draining the mall here, and they say there are over 800 [bodies], a Civil Guard told me...”).

\begin{table}[h!]
\centering
\caption{Linguistic Patterns in TikToks Using LIWC.}
\resizebox{1\columnwidth}{!}{
\begin{tabular}{|l|c|c|c|c|c|}
\hline
\textbf{Category} & \textbf{Mean (0)} & \textbf{Std (0)} & \textbf{Mean (1)} & \textbf{Std (1)} & \textbf{\textit{p}-value} \\ \hline
\textbf{Prepos} & \textbf{10.68} & 3.66 & 8.61 & 3.66 & 0.00001 \\ \hline
\textbf{Health} & \textbf{0.25} & 0.42 & 0.18 & 0.36 & 0.0424 \\ \hline
\textbf{Time} & \textbf{2.56} & 2.74 & 2.01 & 1.64 & 0.0261 \\ \hline
\textbf{Negation} & 1.37 & 1.28 & \textbf{1.93} & 2.58 & 0.0105 \\ \hline
\textbf{Maldec} & 0.11 & 0.30 & \textbf{0.20} & 0.42 & 0.0130 \\ \hline
\textbf{Percept} & 2.02 & 1.36 & \textbf{2.56} & 1.94 & 0.0336 \\ \hline
\end{tabular}
}
\label{tab:tiktok_liwc}
\end{table}

\subsection{Linguistic Patterns in X}

Table~\ref{tab:twitter_liwc} highlights the linguistic patterns observed in tweets. Similar to TikToks, tweets without disinformation also employ a more articulate discourse, using more prepositions (\textit{p} = 0.000001) and negation (\textit{p} = 0.036797).

In tweets spreading disinformation, there is a notable increase in the use of first-person singular pronouns (I). These tweets often cite unverifiable family connections as sources of information (Family), such as “Repito, tengo un familiar directo militar en el parking de Bonaire y están sacando muertos sin parar” (“I repeat, I have a direct family member in the military at the Bonaire parking lot, and they are removing dead bodies non-stop”).

\begin{table}[h!]
\centering
\caption{Linguistic Patterns in X Using LIWC.}
\resizebox{1\columnwidth}{!}{
\begin{tabular}{|l|c|c|c|c|c|}
\hline
\textbf{Category} & \textbf{Mean (0)} & \textbf{Std (0)} & \textbf{Mean (1)} & \textbf{Std (1)} & \textbf{\textit{p}-value} \\ \hline
\textbf{Prepos} & \textbf{14.08} & 5.73 & 11.06 & 6.56 & 0.000001 \\ \hline
\textbf{Negation} & 0.86 & 1.67 & \textbf{1.32} & 2.21 & 0.0368 \\ \hline
\textbf{I} & 0.31 & 1.14 & \textbf{0.54} & 1.42 & 0.0403 \\ \hline
\textbf{Family} & 0.02 & 0.21 & \textbf{0.13} & 0.79 & 0.0320 \\ \hline
\end{tabular}
}
\label{tab:twitter_liwc}
\end{table}

The results indicate clear differences in linguistic patterns between disinformation and trustworthy  content on both platforms. TikToks and tweets without disinformation  tend to use more articulate language with a higher focus on factual information, while disinformation  often incorporates perceptual language, negations, and personal anecdotes to enhance credibility

\uppercase{\section{Audio Features in TikToks}}
\label{sec:audio_analysis}

Audio is a key component of TikTok videos, contributing significantly to their emotional and persuasive impact. In this study, we extracted a wide range of audio features to analyze the role of sound in differentiating disinformation and trustworthy content. The extracted features include:

\begin{itemize}
    \item \textbf{Zero Crossing Rate (ZCR)}. It measures the rate at which the signal changes its sign, reflecting the noisiness or percussive nature of the sound.
    \item \textbf{Root Mean Square Energy (RMS)}. It captures the loudness or energy of the audio signal, often linked to the intensity of the content.
    \item \textbf{Spectral Centroid}. It indicates the center of gravity of the spectrum, associated with the perceived brightness of the audio.
    \item \textbf{Spectral Rolloff}. It represents the frequency below which a certain percentage (e.g., 85\%) of the spectral energy is concentrated, often used to distinguish harmonic and percussive elements.
    \item \textbf{Chroma Features}. They measure the energy distribution among the 12 pitch classes, representing harmonic content.
    \item \textbf{Spectral Bandwidth}. It quantifies the range of frequencies present in the signal, reflecting its tonal richness.
    \item \textbf{Spectral Flatness}. It describes the noisiness of the signal by measuring how flat the power spectrum is.
    \item \textbf{Mel-Frequency Cepstral Coefficients (MFCCs)}. They capture the timbral aspects of the audio, essential for distinguishing speech and music characteristics.
    \item \textbf{Harmonics-to-Noise Ratio (HNR)}. It indicates voice quality and clarity by comparing harmonic and noise components.
    \item \textbf{Tempo}. It estimates the speed of the audio, which can evoke urgency or calmness in the listener.
    \item \textbf{Onset Strength}. It measures the intensity of sound onsets, reflecting rhythmic and dynamic variations.
    \item \textbf{Pitch}. It refers to the perceived fundamental frequency of the signal, often linked to emotional tones such as excitement or seriousness.
    \item \textbf{Spectral Contrast}. It represents differences in amplitude between peaks and valleys in the power spectrum, providing insights into the dynamic range of the audio.
    \item \textbf{Tonnetz Features}. They capture tonal harmonic relationships, offering information about the musical nature of the audio.
\end{itemize}

For each feature, both the \textbf{mean} and \textbf{standard deviation (std)} were calculated to capture the central tendency and variability of the audio properties.\newline

\begin{table}[h!]
\centering
\caption{Significant Audio feature comparison between trustworthy (0) and disinformation (1) groups.}
\small
\begin{tabular}{|p{0.14\columnwidth}|p{0.12\columnwidth}|p{0.12\columnwidth}|p{0.12\columnwidth}|p{0.12\columnwidth}|p{0.10\columnwidth}|}
\hline
\textbf{Feature} & \textbf{Mean (0)} & \textbf{Std (0)} & \textbf{Mean (1)} & \textbf{Std (1)} & \textbf{\textit{p}-value} \\ \hline
\textbf{zcr} & \textbf{0.094} & 0.031 & 0.085 & 0.029 & 0.019 \\ \hline
\textbf{zcr std} & \textbf{0.072} & 0.030 & 0.060 & 0.029 & 0.006 \\ \hline
\textbf{spectral centroid} & \textbf{1902.41} & 465.65 & 1682.81 & 440.07 & 0.0002 \\ \hline
\textbf{spectral centroid std} & \textbf{935.15} & 313.20 & 798.22 & 337.14 & 0.0049 \\ \hline
\textbf{spectral rolloff} & \textbf{3692.10} & 996.91 & 3244.93 & 918.65 & 0.0010 \\ \hline
\textbf{spectral rolloff std} & \textbf{1764.06} & 470.15 & 1570.43 & 583.68 & 0.019 \\ \hline
\textbf{spectral bandwidth} & \textbf{1984.54} & 339.32 & 1799.34 & 348.43 & 0.0001 \\ \hline
\textbf{spectral flatness} & \textbf{0.022} & 0.029 & 0.017 & 0.028 & 0.007 \\ \hline
\textbf{spectral flatness std} & \textbf{0.048} & 0.063 & 0.044 & 0.069 & 0.030 \\ \hline
\textbf{onset strength} & \textbf{1.80} & 0.36 & 1.66 & 0.40 & 0.006 \\ \hline
\textbf{mfcc2 mean} & 111.10 & 25.04 & \textbf{120.40} & 28.08 & 0.022 \\ \hline
\textbf{mfcc5 mean} & \textbf{3.16} & 10.85 & -0.44 & 11.02 & 0.010 \\ \hline
\textbf{mfcc6 std} & \textbf{16.56} & 3.77 & 15.54 & 3.97 & 0.021 \\ \hline
\textbf{contrast7 mean} & 49.03 & 2.60 & \textbf{50.13} & 3.67 & 0.025 \\ \hline
\textbf{contrast6 std} & \textbf{7.58} & 1.55 & 7.15 & 1.47 & 0.015 \\ \hline
\textbf{tonnetz4 mean} & -0.001 & 0.051 & \textbf{0.022} & 0.074 & 0.003 \\ \hline
\end{tabular}
\label{tab:audio_features}
\end{table}

The analysis revealed significant differences in audio features between trustworthy (0) and disinformation (1) groups, as shown in Table~\ref{tab:audio_features}, reflecting distinct acoustic and tonal patterns:

\begin{itemize}
    \item \textbf{Zero Crossing Rate (ZCR, ZCR\_STD).} 
    Trustworthy audios exhibited higher values, indicating more frequent transitions across the zero axis in the waveform. This reflects more energetic and articulated speech, characteristic of clear and direct communication.

    \item \textbf{Spectral Features:}
    \begin{itemize}
        \item \textit{Spectral Centroid.} Higher in trustworthy content, representing brighter tones with more emphasis on higher frequencies.
        \item \textit{Spectral Bandwidth.} Wider in trustworthy content, indicating a broader range of frequencies around the centroid.
        \item \textit{Spectral Rolloff.} Higher in trustworthy posts, reflecting stronger energy in higher frequencies, which enhances tonal clarity.
    \end{itemize}

    \item \textbf{Spectral Flatness.}. Through the visualization of videos with high spectral flatness values, we observed that they frequently featured narrators delivering information in a neutral, monotone, or robotic style, often associated with objective news reporting about tragedies or serious events. This explains why spectral flatness was higher in trustworthy audios. In contrast, disinformation audios displayed lower spectral flatness, indicating more tonal content and greater variation. These variations are likely aimed at dramatizing events, evoking emotional responses, and manipulating the listener’s perception.
    
    \item \textbf{Onset Strength.} Higher onset strength was observed in trustworthy audios, reflecting strong and clear acoustic events such as sharp changes in amplitude or the beginning of words and sounds. In contrast, lower onset strength in disinformation audios suggests smoother transitions and less dynamic emphasis, which might be used to create a more conversational or emotionally manipulative tone.

    \item \textbf{MFCCs (Mel-Frequency Cepstral Coefficients):}
    \begin{itemize}
        \item \textit{MFCC2.} Higher in disinformation, capturing lower frequency tonal characteristics, associated with deeper and emotionally charged tones.
        \item \textit{MFCC5 and MFCC6.} Higher in trustworthy content, indicating richer and clearer tonal qualities in the mid-frequency range, aligned with neutral and structured speech.
    \end{itemize}

    \item \textbf{Spectral Contrast (Contrast7).} Higher in disinformation, particularly in the highest frequency bands, helping dramatize content and attract attention.

    \item \textbf{Tonnetz (Tonnetz4).} Videos with high Tonnetz values often included eerie or mysterious music. Observational analysis of such videos revealed their use in disinformation to create a suspenseful or manipulative tone, amplifying the emotional impact.
\end{itemize}
 
Trustworthy audios are characterized by brighter tones, dynamic events, and monotone or robotic narration styles, optimized for clarity and credibility. Disinformation audios, on the other hand, emphasize tonal variation, emotional depth, and harmonic complexity, often accompanied by manipulative musical elements designed to captivate and influence the listener.

\section{\uppercase{Models}}
\label{sec:models}
To train the models for disinformation  detection, tweets and transcriptions of TikToks were combined into a unified dataset. To minimize bias, hashtags and user mentions were removed from the text to prevent the models from relying on these features for predictions. Given the limited size of the dataset, we employed three main approaches: a traditional machine learning model, fine-tuning of transformer-based models, and a GPT-4o Few-Shot learning approach.

The models used were \textbf{roberta-base-bne}\footnote{\url{https://huggingface.co/PlanTL-GOB-ES/roberta-base-bne}}, \textbf{roberta-large-bne}\footnote{\url{https://huggingface.co/PlanTL-GOB-ES/roberta-large-bne}}, \textbf{xlm-roberta-large-twitter}\footnote{\url{https://huggingface.co/sdadas/xlm-roberta-large-twitter}}, \textbf{longformer-base-4096-bne-es}\footnote{\url{https://huggingface.co/PlanTL-GOB-ES/longformer-base-4096-bne-es}}, \textbf{fake-news-detection-spanish}\footnote{\url{https://huggingface.co/Narrativaai/fake-news-detection-spanish}}, \textbf{SVM+TF-IDF}, and \textbf{GPT-4o Few-Shot}. We also experimented with fine-tuning \textbf{roberta-large-bne} by incorporating \textbf{audio features} from TikTok. These features were normalized and combined with text embeddings from the model's [CLS] token. The fused representations were passed through a classifier, leveraging both textual and audio information to enhance disinformation detection in a multimodal context.

To ensure robust evaluation, 5-fold stratified cross-validation was employed for the traditional and transformer-based models, maintaining the class distribution in each fold. This approach provided a comprehensive assessment of model performance, minimizing the risk of overfitting and ensuring reliable generalization to unseen data. For the GPT-4o Few-Shot approach, a single evaluation was conducted.

The performance of each model was evaluated using the following metrics: Accuracy, F1-Score, Precision, and Recall. The results are summarized in Table \ref{tab:results}, where the best-performing model for each metric is highlighted in \textbf{bold}.

\begin{table*}[ht]
\centering
\caption{Performance Metrics of Different Models for Disinformation Detection}
\label{tab:results}
\resizebox{\textwidth}{!}{%
\begin{tabular}{lcccc}
\toprule
\textbf{Model} & \textbf{Accuracy} & \textbf{F1-Score} & \textbf{Precision} & \textbf{Recall} \\
\midrule
\textbf{roberta-base-bne} & $0.7497 \pm 0.0506$ & $0.7565 \pm 0.0575$ & $0.7668 \pm 0.0313$ & $0.7486 \pm 0.0839$ \\
\textbf{roberta-large-bne} & $0.7512 \pm 0.0177$ & $0.7520 \pm 0.0270$ & $0.7885 \pm 0.0248$ & $0.7220 \pm 0.0543$ \\
\textbf{roberta-large-bne + audio features} & $0.7646 \pm 0.0292$ & $0.7664 \pm 0.0272$ & $0.7726 \pm 0.0432$ & $0.7634 \pm 0.0484$ \\
\textbf{xlm-roberta-large-twitter} & $0.7274 \pm 0.0175$ & $0.7358 \pm 0.0192$ & $0.7505 \pm 0.0167$ & $0.7229 \pm 0.0352$ \\
\textbf{longformer-base-4096-bne-es} & $0.7266 \pm 0.0473$ & $0.7035 \pm 0.0681$ & $0.8069 \pm 0.0361$ & $0.6285 \pm 0.0943$ \\
\textbf{fake-news-detection-spanish} & $0.7298 \pm 0.0454$ & $0.7377 \pm 0.0416$ & $0.7541 \pm 0.0486$ & $0.7222 \pm 0.0370$ \\
\textbf{SVM+TF-IDF} & $0.7604 \pm 0.0173$ & $\textbf{0.7949} \pm 0.0133$ & $0.7227 \pm 0.0165$ & $\textbf{0.8831} \pm 0.0087$ \\
\textbf{GPT-4o Few-Shot} & $\textbf{0.7834}$ & $0.7867$ & $\textbf{0.8150}$ & $0.7602$ \\
\bottomrule
\end{tabular}%
}
\end{table*}

The \textit{SVM+TF-IDF} model demonstrated the highest effectiveness in terms of \textit{F1-Score} and \textit{Recall}, achieving scores of $0.7949$ and $0.8831$, respectively. This superiority can be attributed to the Support Vector Machine's ability to efficiently handle high-dimensional feature spaces even with small datasets, thereby avoiding overfitting and providing robust generalization. The simplicity and efficiency of traditional models like SVM make them particularly suitable when data resources are limited.

On the other hand, transformer-based models such as \textit{roberta-base-bne} and \textit{roberta-large-bne} exhibited competitive performance, albeit slightly inferior to SVM in this scenario of limited data. However, it is important to note that these models possess significant potential for scalability and improved performance when larger training datasets are available.

Notably, by incorporating audio features into \textit{roberta-large-bne}, we observed a significant improvement in both \textit{Accuracy} and \textit{F1-Score}, achieving $0.7646$ in accuracy (an increase of 1.34 percentage points over the standard \textit{roberta-large-bne}) and $0.7664$ in F1-Score (an increase of 1.44 percentage points). Furthermore, this multimodal approach outperformed the \textit{SVM+TF-IDF} model in terms of \textit{Accuracy}, demonstrating the potential of combining textual and audio features to enhance performance in disinformation detection tasks.

Finally, the Few-Shot approach using \textbf{GPT-4o} achieved the highest accuracy of $0.7834$, demonstrating the effectiveness of large-scale language models when provided with well-designed and specific prompts. This result highlights the potential of LLMs for automated annotation and classification tasks, even without domain-specific training.

\section{\uppercase{Conclusions and Future Work}}
\label{sec:conclusions}
\subsection{Conclusions}

This study analyzed the spread of disinformation on TikTok and X during the DANA event in Valencia, Spain, which was marked by severe flooding. A dataset of 650 posts was manually annotated to distinguish between disinformation and trustworthy content. Additionally, a separate experiment was conducted using GPT-4o in a Few-Shot setup to evaluate its potential as a tool for automated annotation. GPT-4o demonstrated substantial agreement with the manual labels (Cohen's kappa of 0.684 for four categories and 0.695 for two categories).

Emotion analysis revealed that disinformation on X is primarily associated with heightened sadness and fear, while on TikTok, it correlates with increased anger and disgust. The identification of this divergence in predominant emotions across different social networks is a distinctive result of our study. First, our work reinforces the conclusion, already reached by many previous studies, that appealing to emotions is a recurring strategy in deliberately misleading messages \cite{osmundsen2021partisan, mcloughlin2024misinformation}. However, thanks to the comparative approach of our study—where we analyzed the prevailing emotional traits in posts on X and TikTok—we have identified a phenomenon that has been scarcely investigated until now: the fact that, on each platform, disinformation content operates under a different emotional paradigm. This conclusion underscores the need to further understand the impact of emotional dimensions on credulity toward disinformation content, as well as the predisposition to share it \cite{pennycook2021psychology}. Linguistic analysis using the LIWC dictionary indicated that trustworthy content tends to use more articulate and factual language, whereas disinformation employs negations, perceptual words, and personal anecdotes to enhance credibility.

The analysis of TikTok audio features revealed distinct patterns: trustworthy audio showed brighter tones, stronger onset strength, and higher spectral flatness, reflecting clarity and neutrality, while disinformation audio used tonal variations and deeper MFCCs to evoke emotions and manipulate perceptions. These findings highlight the value of audio analysis in multimodal disinformation detection.

In terms of detection models, the traditional SVM+TF-IDF model outperformed transformer-based models and the GPT-4o Few-Shot approach in \textit{F1-Score} and \textit{Recall}, benefiting from the limited dataset size. However, transformer-based models showed competitive performance and potential for scalability with larger datasets. Notably, by incorporating audio features into \textit{roberta-large-bne}, we observed an improvement in both \textit{Accuracy} and \textit{F1-Score} compared to its text-only counterpart, as well as a higher accuracy than the SVM+TF-IDF model. This demonstrates the value of multimodal approaches in disinformation detection tasks. The GPT-4o Few-Shot model demonstrated high accuracy, highlighting the effectiveness of LLMs in automated classification tasks without extensive domain-specific training.

\subsection{Future Work}

Future research can expand on this study by:

\begin{itemize}
    \item \textbf{Increasing Dataset Size}. Incorporate more tweets and TikToks, as well as additional social media platforms.
    \item \textbf{Multimodal Analysis}. Include images and videos to capture a broader range of disinformation strategies.
    \item \textbf{Emotional Divergences}. Examine how distinct emotional triggers influence disinformation dissemination on different social media platforms.

    \item \textbf{Advanced Models}. Explore sophisticated transformer architectures and ensemble methods for enhanced performance.
    \item \textbf{Cross-Linguistic Studies}. Analyze disinformation patterns across different languages and regions.
    \item \textbf{Network Analysis}. Investigate user behaviors and network structures that facilitate disinformation spread.
    \item \textbf{Refining Few-Shot Techniques}. Improve prompt designs and methodologies for large language models to reduce reliance on manual annotations.
    \item \textbf{Intervention Strategies}. Develop effective counter-disinformation measures based on identified emotional and linguistic patterns.
    \item \textbf{Intelligent Agents}. Design and implement intelligent agents capable of real-time detection and mitigation of disinformation. These agents could combine machine learning techniques with rule-based systems to analyze linguistic and emotional cues, detect malicious content, and take automated countermeasures. For instance, intelligent agents could:
    \begin{itemize}
        \item Flag potential disinformation for review or further analysis.
        \item Provide users with context or verified information to counter false claims.
        \item Interact with social media algorithms to limit the spread of harmful content while promoting verified, accurate information.
    \end{itemize}
\end{itemize}

Addressing these areas will enhance the detection and mitigation of disinformation on social media, thereby strengthening the integrity of information ecosystems during critical events.

\section*{Acknowledgements}
This work was done in the framework of the Iberian Digital Media Observatory (IBERIFIER Plus), co-funded by the EC under the Call DIGITAL-2023-DEPLOY-04 (Grant 101158511), and of the
Malicious Actors Profiling and Detection in Online Social Networks Through Artificial Intelligence (MARTINI) research project, funded by MCIN/AEI/10.13039/501100011033 and by NextGenerationEU/PRTR (Grant PCI2022-135008-2).

\bibliographystyle{apalike}
{\small
\bibliography{example}}

\appendix

\section{Detailed Few-Shot Annotation Prompt for GPT-4o}
\label{sec:appendix_prompt}

The following prompt was originally provided in Spanish to align with the language of the data but for the sake of comprehension for non-Spanish speakers, it has been translated into English in this article:

\textbf{Context:}
The Depresión Aislada en Niveles Altos (DANA) that affected Valencia and other areas of Spain in October 2024 caused severe flooding and the spread of disinformation. Common types of disinformation included:

\begin{itemize}
    \item \textbf{False claims about dam openings or breaks}, such as Forata.
    \item \textbf{Conspiracy theories about climate manipulation} (e.g., HAARP, Morocco, geoengineering).
    \item \textbf{Rumors of hidden deaths or transport in refrigerated trucks} (e.g., parking at Bonaire).
    \item \textbf{False accusations of rejected aid or mismanagement of donations}, such as delays or destruction of aid by the Red Cross.
    \item \textbf{Exaggerated numbers of victims, floods, or negligence}.
    \item \textbf{Alarmist content about basic services} (e.g., water supply, healthcare disruptions).
    \item \textbf{Unfounded accusations against political figures or governments}.
    \item \textbf{Claims of Valencia's meteorological radar being non-functional} (when it was operational).
    \item \textbf{Rumors of censorship or media manipulation} by public figures (e.g., YouTube or Iker Jiménez).
\end{itemize}

\textbf{Labeling Guidelines:}

\begin{itemize}
    \item \textbf{0 (Not related to DANA):} Tweets unrelated to DANA, even indirectly. \\
    \textbf{Example:} ``Just had the best paella in Valencia. Highly recommend!''
    \item \textbf{1 (Related to DANA, but not disinformation):} Tweets discussing real events or consequences of DANA without spreading disinformation. \\
    \textbf{Example:} ``Thousands of volunteers are helping victims of DANA in Valencia.''
    \item \textbf{2 (Disinformation mentioned, but criticized):} Tweets mentioning disinformation about DANA to criticize or report it, including posts debunking or ridiculing falsehoods. \\
    \textbf{Example:} ``Don’t believe the fake news about hidden deaths in Bonaire; it has been debunked by official sources.''
    \item \textbf{3 (Disinformation):} Tweets actively spreading false information, rumors, or conspiracy theories about DANA, or unfounded accusations against political figures. \\
    \textbf{Example:} ``They’re hiding the truth! Over 800 bodies were found in the Bonaire parking lot, and no one is reporting it.''
\end{itemize}

Tweet: [TWEET HERE]\newline Return only the number:

\end{document}